\documentclass[journal]{IEEEtran}
\usepackage{graphicx}
\usepackage{verbatim}
\usepackage{latexsym}
\usepackage{algorithmic}

\begin{document}
\title{Multi-vehicle Cooperative Control Using 
       Mixed Integer Linear Programming}
\author{Matthew~G.~Earl and~Raffaello~D'Andrea%
\thanks{M. G. Earl (corresponding author), Theoretical and
Applied Mechanics Department, Cornell University, Ithaca, NY 14853,
(e-mail: {\tt mge1@cornell.edu}).}
\thanks{R. D'Andrea, Mechanical and Aerospace Engineering 
Department, 101 Rhodes Hall, Cornell University, Ithaca, NY 14853, 
(e-mail: {\tt rd28@cornell.edu}).}}

\maketitle
\begin{abstract}
We present methods to synthesize cooperative strategies for
multi-vehicle control problems using mixed integer linear programming.  
Complex multi-vehicle control problems are expressed
as mixed logical dynamical systems.  Optimal strategies for these
systems are then solved for using mixed integer linear programming.
We motivate the methods on problems derived from an adversarial game
between two teams of robots called RoboFlag.  We assume the strategy
for one team is fixed and governed by state machines. The strategy for
the other team is generated using our methods. Finally, we perform an
average case computational complexity study on our approach.
\end{abstract}


\IEEEpeerreviewmaketitle

\section{Introduction}
For many problems, a team of vehicles can accomplish an
objective more efficiently and more effectively than a single vehicle
can.  Some examples include target intercept~\cite{Beard02},
search~\cite{Beard03b}, terrain mapping~\cite{Marco03}, object 
manipulation~\cite{Sugar02}, surveillance, and  space-based interferometry.
For these problems, it is desirable to design a multi-vehicle
cooperative control strategy.

There is a large literature on cooperative control.  Work from team
theory~\cite{Marschak72,Ho72} considers the case where team members
have different information and the objective function is quadratic.
Cooperative estimation for reconnaissance problems is considered
in~\cite{Ous04a}.  In~\cite{Richards02,Bellingham02,Earl02}  mixed
integer linear programming is used for multi-vehicle target assignment
and intercept problems.  Hierarchical methods are used for cooperative
rendezvous in~\cite{McLain01} and for target assignment and intercept
in~\cite{Beard02}.  A review from the machine learning perspective is
presented in~\cite{Stone00}.  There are several recent compilations of
cooperative control articles in~\cite{Arkin97,Balch00,Murphey02}.

In this paper, we propose a hybrid systems approach for modeling and
cooperative control of multi-vehicle systems.  We use the class of
hybrid systems called mixed logical dynamical
systems~\cite{Bemporad00}, which are governed by difference equations
and logical rules and are subject to linear inequality constraints.
The main motivation for using mixed logical dynamical systems is their
ability to model a wide variety of multi-vehicle problems and the ease
of modifying problem formulations.  In our approach, a problem is
modeled as a mixed logical dynamical system, which we represent as a
mixed integer linear program (MILP). Then, to generate a cooperative
control strategy for the system, the MILP is solved using
AMPL~\cite{fourer93} and CPLEX~\cite{ilog00}.

Posing a multi-vehicle control problem in a MILP framework involves
modeling the vehicle dynamics and constraints, modeling the
environment, and expressing the objective.  To demonstrate the
modeling procedure and our approach, we consider control problems
involving Cornell's multi-vehicle system called RoboFlag.  For an
introduction to RoboFlag, see the papers from the invited session on
RoboFlag in the Proceedings of the 2003 American Control
Conference~\cite{campbell03,d'andrea03a,d'andrea03b}.  

Our focus is to find optimal solutions using a flexible methodology,
which is why we use MILP.  However, because MILP is in the NP-hard
computation class~\cite{Garey}, the methods may not be fast enough for
real-time control of systems with large problem formulations. In this
case, the methods can be used to explore optimal cooperative behavior
and as benchmarks to evaluate the performance of heuristic or
approximate methods.  In Section~\ref{sec:discussion}, we discuss
several methods for reducing the computational requirements of our
MILP approach so that it can be more readily used in real-time
applications. 

Our approach for multi-vehicle control, first presented
in~\cite{Earl02b,Earl02}, was developed independently from a similar
approach developed by Richards et.~al.~\cite{Richards02b}.  Next, we
list some of the noteworthy aspects of our approach.  First, the
environment which we demonstrate our methods involves an adversarial
component. We model the intelligence of the adversaries with state
machines.  Second, our approach allows multiple, possibly nonuniform,
time discretizations.  Discretizing continuous variables in time is
necessary for MILP formulations.  Using many time steps results in
large MILPs that require a considerable amount of computation time to
solve.  Support for nonuniform discretizations in time allows the use
of intelligent time step selection algorithms for the generation of
more efficient MILP problem formulations~\cite{Earl04b}.  Finally,
because we include the vehicle dynamics in the problem formulation,
the resulting trajectories are feasible, which is advantageous because
they can be applied directly to the multi-vehicle system. In order to
express the vehicle dynamics efficiently in our MILP formulation, we
restrict the control input to each vehicle in a way that allows
near-optimal performance, as presented in~\cite{Nagy04}.

The paper is organized as follows:  First, we consider vehicle problems
that have relatively simple formulations. Then we add features in each
section until we arrive at the RoboFlag multi-vehicle problems.  In
Section~\ref{vehicle_dynamics}, we introduce the dynamics of the
vehicles used to motivate our approach, and we formulate and solve a
single vehicle minimum control effort trajectory generation problem.
We build upon this in Section~\ref{sec:avoidance} adding obstacles
that must be avoided.  In Section~\ref{sec:dd}, we show how to
generate optimal team strategies for RoboFlag problems.  In
Section~\ref{sec:ctime}, we perform an average case computational
complexity study on our approach.  Finally, in
Section~\ref{sec:discussion}, we discuss our methods and ways in which
they can be be applied.  All files for generating the plots found in
this paper are available online~\cite{EarlWebPage}. 

\section{Vehicle dynamics}
\label{vehicle_dynamics}
Multi-vehicle control problems involving the wheeled robots of
Cornell's RoboCup Team~\cite{d'andrea01,stone01} are used to motivate
the methods presented in this paper.  In this section, we show how to
simplify their nonlinear governing equations using a procedure
from~\cite{Nagy04}. The result is a linear set of governing equations
coupled by a nonlinear constraint on the control input, which admits
feasible vehicle trajectories and allows near-optimal performance. We
then show how to represent the simplified system in a MILP problem
formulation.

Each vehicle has a three-motor omni-directional drive,
which allows it to  move along any direction irrespective of its
orientation.  
The nondimensional
governing equations of each vehicle are given by
\begin{equation}
  \left[ \begin{array}{c}
    \ddot{x}(t) \\
    \ddot{y}(t) \\
    \ddot{\theta}(t) 
  \end{array} \right] +
  \left[ \begin{array}{c}
    \dot{x}(t) \\
    \dot{y}(t) \\
    \frac{2mL^2}{I}\dot{\theta}(t) 
  \end{array} \right] =
  \mathbf{u}(\theta(t),t),
\end{equation}
where $\mathbf{u}(\theta(t),t) = \mathbf{P}(\theta(t)) \mathbf{U}(t)$,
\begin{equation}
  \mathbf{P}(\theta) = 
  \left[ \begin{array}{ccc}
    -\sin(\theta)&-\sin(\frac{\pi}{3}-\theta)&\sin(\frac{\pi}{3}+\theta) \\ 
    \cos(\theta)&-\cos(\frac{\pi}{3}-\theta)&-\cos(\frac{\pi}{3}+\theta) \\ 
    1 & 1 & 1 
  \end{array} \right],
\end{equation}
and $\mathbf{U}(t) = (U_x(t), U_y(t), U_z(t)) \in \mathcal{U}$.
In these equations, $(x(t),y(t))$ are the coordinates of the vehicle on
the playing field, $\theta(t)$ is the orientation of the vehicle,
$\mathbf{u}(\theta(t),t)$ is the $\theta(t)$-dependent
control input, $m$ is the mass of the vehicle, $I$
is the vehicle's moment of inertia, $L$ is the distance from the drive
to the center of mass, and $U_i(t)$ is the voltage applied to motor
$i$.  The set of admissible voltages $\mathcal{U}$ is given by the
unit cube, and the set of admissible control inputs is given by
$P(\theta) \mathcal{U}$.

These governing equations are coupled and nonlinear.  To
simplify them, we replace the set $P(\theta)\mathcal{U}$ with the
maximal $\theta$-independent set found by taking the intersection of
all possible sets of admissible controls.  The result is a
$\theta$-independent control input, denoted $(u_x(t), u_y(t),
u_z(t))$, that is subject to the inequality constraints
$u_x(t)^2 + u_y(t)^2 \leq (3-|u_\theta(t)|)^2/4$
and $|u_\theta(t)| \leq 3$.

Using the restricted set as the set of allowable control inputs, the 
governing equations decouple and are given by 
\begin{equation}
  \left[ \begin{array}{c}
    \ddot{x}(t) \\
    \ddot{y}(t) \\
    \ddot{\theta}(t) 
  \end{array} \right] +
  \left[ \begin{array}{c}
    \dot{x}(t) \\
    \dot{y}(t) \\
    \frac{2mL^2}{I}\dot{\theta}(t) 
  \end{array} \right] =
  \left[ \begin{array}{c}
    u_x(t) \\
    u_y(t) \\
    u_\theta(t) 
  \end{array} \right].
\end{equation}
The constraints on the control input couple the degrees of freedom.  

To decouple the $\theta$ dynamics, we set $|u_\theta(t)| \leq 1$. Then
the constraint on the control input becomes
\begin{eqnarray}
\label{nl_u_constraint}
&&u_x(t)^2 + u_y(t)^2 \leq 1.
\end{eqnarray}
Now the equations of motion for the translational dynamics of the
vehicle are given by 
\begin{eqnarray}
\label{eqn:gov}
&&\ddot{x}(t) + \dot{x}(t) = u_x(t),\nonumber\\
&&\ddot{y}(t) + \dot{y}(t) = u_y(t),
\end{eqnarray}
subject to equation~(\ref{nl_u_constraint}).
In state space form, equation~(\ref{eqn:gov}) is
$\dot{\mathbf{x}}(t) = \mathbf{A}_c \mathbf{x}(t) + \mathbf{B}_c
\mathbf{u}(t)$,
where $\mathbf{x} = (x,y,\dot{x},\dot{y})$ is the state and
$\mathbf{u} = (u_x,u_y)$ is the control input.

By restricting the admissible control inputs we have simplified the
governing equations in a way that allows near optimal performance as
shown in~\cite{Nagy04}.  This procedure allows real-time calculation
of many near-optimal trajectories and has been successfully used by
Cornell's RoboCup team~\cite{d'andrea01,stone01,Nagy04}. 

To represent the governing equations in a MILP framework, we discretize
the control input in time and require it be constant between time steps.
The result is a set of linear discrete time governing equations.

Let $N_u$ be the number of discretization steps for the control input
$\mathbf{u}(t)$, let $t_u[k]$ be the
time at step $k$, and let $T_u[k]>0$ be the time between steps $k$ and
$k+1$, for $k \in \{0,\ldots,N_u-1 \}$. 
The discrete time governing equations are given by

\begin{equation}
  \mathbf{x}_u[k+1] = \mathbf{A}[k] \mathbf{x}_u[k] 
  + \mathbf{B}[k] \mathbf{u}[k],
  \label{disdyn}
\end{equation}
where $\mathbf{x}_u[k] = \mathbf{x}(t_u[k])$, 
$\mathbf{u}[k] = \mathbf{u}(t_u[k])$, 
\begin{eqnarray}
  \mathbf{A}[k] =  
  \left[ \begin{array}{cccc}
    1 & 0 & 1-e^{-T_u[k]} & 0 \\ 
    0 & 1 & 0 & 1-e^{-T_u[k]} \\ 
    0 & 0 & e^{-T_u[k]} & 0 \\ 
    0 & 0 & 0 & e^{-T_u[k]}  
  \end{array} \right],
  \nonumber
\end{eqnarray}
\begin{eqnarray}
  \mathbf{B}[k] =  
  \left[ \begin{array}{cc}
    T_u[k]-1+e^{-T_u[k]} & 0 \\ 
    0 & T_u[k]-1+e^{-T_u[k]} \\ 
    1-e^{-T_u[k]} & 0 \\ 
    0 & 1-e^{-T_u[k]}  
  \end{array} \right],
  \nonumber
\end{eqnarray}
$\mathbf{x}_u[k] = (x_u[k],y_u[k],\dot{x}_u[k],\dot{y}_u[k])$, and
$\mathbf{u}[k] = (u_{x}[k],u_{y}[k])$.  The coefficients
$\mathbf{A}[k]$ and $\mathbf{B}[k]$ are functions of $k$ because we
have allowed for nonuniform time discretizations.  Because there will
be several different time discretizations used in this paper, we use
subscripts to differentiate them.  In this section, we use the
subscript $u$ to denote variables associated with the discretization
in the control input $\mathbf{u}(t)$.  

The discrete time governing equations can be solved explicitly to obtain 
\setlength{\arraycolsep}{0.0em}
\begin{eqnarray}
  x_u[k]&{}={}&x_u[0] +
  \left(1-e^{-t_u[k]}\right)\dot{x}_u[0]
  \nonumber\\
  &&{+}\:\sum_{i=0}^{k-2} 
  \left(
    \left(T_u[i] - 1 + e^{-T_u[i]}\right) u_{x}[i] 
  \right.\nonumber\\
  &&{+}\:\left.\left(1-e^{t_u[i+1] - t_u[k]}\right)
  \left(1-e^{-T_u[i]}\right) u_{x}[i]
  \right)\nonumber\\
  &&{+}\:\left(T_u[k-1]-1+e^{-T_u[k-1]}\right)u_{x}[k-1],\nonumber
\end{eqnarray}
\setlength{\arraycolsep}{5pt}%
\setlength{\arraycolsep}{0.0em}
\begin{eqnarray}
  \dot{x}_u[k]&{}={}&e^{-t_u[k]} \dot{x}_u[0]\nonumber\\
  &&{+}\:\sum_{i=0}^{k-2} \left(e^{t_u[i+1]-t_u[k]}
  \left(1-e^{-T_u[i]}\right)u_{x}[i]\right)\nonumber\\
  &&{+}\:\left(1-e^{-T_u[k-1]}\right)u_{x}[k-1],\nonumber 
\end{eqnarray}
\setlength{\arraycolsep}{5pt}%
and similarly for $y_u[k]$ and $\dot{y}_u[k]$.  

In later sections of this paper it will be necessary to represent the
position of the vehicle at times between control discretization
steps, in terms of the control input.  Because the set of governing
equations is linear, given the discrete state $\mathbf{x}_u[k]$ and
the control input $\mathbf{u}[k]$, we can calculate the vehicle's
state at any time $t$ using the following equations:
\setlength{\arraycolsep}{0.0em}
\begin{eqnarray}
  x(t)&{}={}& x_u[k] + (1 - e^{t_u[k] - t}) \dot{x}_u[k]\nonumber\\
  &&{+}\:(t - t_u[k] - 1 + e^{t_u[k] - t}) u_{x}[k],\nonumber\\
  \dot{x}(t)&{}={}& (e^{t_u[k] - t}) \dot{x}_u[k]
  + (1 - e^{t_u[k] - t})  u_{x}[k],
  \label{inbetween}
\end{eqnarray}
\setlength{\arraycolsep}{5pt}%
where $k$ satisfies $t_u[k] \leq t \leq t_u[k+1]$.  If the time
discretization of the control input is uniform, $T_u[k_u] = T_u$ for
all $k_u$, then $k_u = \lfloor t/T_u \rfloor$. The other components of the
vehicle's state, $y(t)$ and $\dot{y}(t)$, can be calculated in a
similar way.

The control input constraint given by equation~(\ref{nl_u_constraint})
cannot be expressed in a MILP framework because it is nonlinear.  To
incorporate this constraint we approximate it with a set of linear
inequalities that define a polygon.  The polygon
inscribes the region defined by the nonlinear constraint.  We take the
conservative inscribing polygon to guarantee that the set of allowable
controls, defined by the region, is feasible.  
We define the polygon by the set of $M_u$
linear inequality constraints 
\begin{eqnarray}
  &&u_{x}[k] \sin \frac{2 \pi m}{M_u} + 
  u_{y}[k] \cos \frac{2 \pi m}{M_u} \leq \cos \frac{\pi}{M_u}\nonumber\\ 
  &&\forall m \in \{ 1,\ldots, M_u \},
  \label{linconstraint}
\end{eqnarray}
for each step $k \in \{ 1,\ldots,N_u \}$.

To illustrate the approach, we consider the following minimum control
effort trajectory generation problem. Given a
vehicle governed by equations~(\ref{disdyn})
and~(\ref{linconstraint}), find the sequence of control inputs
$\{\mathbf{u}[k]\}_{k=0}^{N_u-1}$ that transfers the vehicle from
starting state $\mathbf{x}(0) = \mathbf{x}_s$ to finishing state
$\mathbf{x}(t_f) = \mathbf{x}_f$ and minimizes the cost function
\setlength{\arraycolsep}{0.0em}
\begin{eqnarray}
J &{}={}& \sum_{k=0}^{N_u-1} \left( |u_{x}[k]| + |u_{y}[k]| \right).
\label{mincontrolcost1}
\end{eqnarray}
\setlength{\arraycolsep}{5pt}%

To convert the absolute values in the cost function to linear form, we
introduce auxiliary continuous variables $z_x[k]$ and $z_y[k]$
and the inequality constraints
\setlength{\arraycolsep}{0.0em}
\begin{eqnarray}
&&-z_x[k] \leq u_{x}[k] \leq z_x[k]\nonumber\\
&&-z_y[k] \leq u_{y}[k] \leq z_y[k].
\label{slackconstraints}
\end{eqnarray}
\setlength{\arraycolsep}{5pt}%
Minimizing $z_x[k]$ subject to the inequalities
$u_{x}[k] \leq z_x[k]$ and
$u_{x}[k] \geq -z_x[k]$ is equivalent to minimizing $|u_x[k]|$
(similarly for $|u_y[k]|)~\cite{Bertsimas97}$.
Using the auxiliary variables, the cost
function can be written as a linear function,
\setlength{\arraycolsep}{0.0em}
\begin{eqnarray}
J &{}={}& \sum_{k=0}^{N_u-1} \left( z_x[k] + z_y[k] \right).
\label{mincontrolcost2}
\end{eqnarray}
\setlength{\arraycolsep}{5pt}%

\begin{figure}
\centering
\includegraphics[height=200pt]{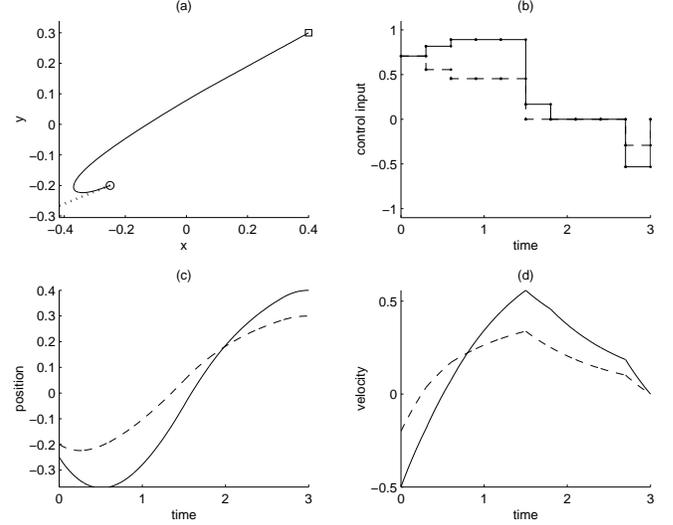}
\caption{Plots of the minimum control effort example.
Figure~(a) shows the vehicle trajectory in the $(x,y)$ plane.  The
circle and dotted line denote the initial position and velocity,
respectively. The square denotes the final position.  Figures~(b)--(d) 
show the time histories of the control inputs, the positions, and
the velocities, respectively.  The solid lines in Figures~(b)--(d)
represent $x$ components and the dotted lines represent $y$
components. The values for the parameters are $N_u = 10$, $M_u =
20$, $T_u[k] = 0.3$ for all $k$, 
$(x_0,y_0,\dot{x}_0,\dot{y}_0) = (-0.25, -0.2, -0.5, -0.2)$, and
$(x_f,y_f,\dot{x}_f,\dot{y}_f) = (0.4, 0.3, 0.0, 0.0)$.
} 
\label{mincontrol1}
\end{figure}

The resulting optimization problem (minimize~(\ref{mincontrolcost2})
subject
to~(\ref{disdyn}),~(\ref{linconstraint}),~(\ref{slackconstraints}),
and the boundary conditions) is in MILP form.  Because binary
variables do not appear in the problem formulation, it is a linear
program and is easily solved to obtain the optimal sequence of control
inputs.  The solution for an example instance is shown in
Figure~\ref{mincontrol1}.

\section{Obstacle avoidance}
\label{sec:avoidance}
In vehicle control, it is necessary to avoid other vehicles,
stationary and moving obstacles, and restricted regions.  In this
section, we show how to formulate and solve these avoidance problems
using MILP.  We start by showing a MILP method to guarantee circular
obstacle avoidance at $N_o$ discrete times. The version of this method
developed in~\cite{Richards02c}, and a similar version developed
independently in in~\cite{Earl02b,Earl02}, uniformly distributes
obstacle avoidance times. Here we present a version of the method that
allows nonuniform distributions of obstacle avoidance times.  

The subscript $o$ is used to
denote variables associated with the time discretization for obstacle
avoidance.  For step $k$, taken to be an element of the set
$\{1,\ldots,N_o\}$, let $t_o[k]$ be the time at which obstacle
avoidance is enforced.  Let $R_{obst}$ denote the radius of the
obstacle. Let $(x_{obst}[k],y_{obst}[k])$ denote the coordinates of
its center at time $t_o[k]$.  We approximate the obstacle with a
polygon, denoted $\mathcal{O}[k]$, defined by a set of $M_o$
inequalities.  The polygon is given by
\setlength{\arraycolsep}{0.0em}
\begin{eqnarray} 
  \mathcal{O}[k]&{}:={}&\{ \mbox{ } (\bar{x},\bar{y}) :\nonumber\\
  &&(\bar{x}-x_{obst}[k]) \sin \frac{2 \pi m}{M_{o}}\nonumber\\
  &&{+}\:(\bar{y}-y_{obst}[k]) \cos \frac{2 \pi m}{M_{o}} 
    \leq R_{obst}\nonumber\\
  &&\forall m \in \{1,\ldots,M_{o}\} \mbox{ } \}.
  \label{obst_con1}
\end{eqnarray}
\setlength{\arraycolsep}{5pt}%

To guarantee obstacle avoidance at time $t_o[k]$, the coordinates of
the vehicle must be outside the region $\mathcal{O}[k]$. This
avoidance condition can be written as $(x_o[k],y_o[k]) \notin
\mathcal{O}[k]$, where $(x_o[k],y_o[k])$ are the coordinates of the
vehicle at time $t_o[k]$.  Here $x_o[k] = x(t_o[k])$ and
$y_o[k]=y(t_o[k])$ are expressed in terms of the control inputs using
equation~(\ref{inbetween}).

Because at least one constraint defining the region $\mathcal{O}[k]$
must be violated in order to avoid the obstacle, the avoidance
condition is equivalent to the following condition: 
\begin{eqnarray}
\label{avoidconstraints3}
  &&\mbox{there exists an } m \mbox{ such that}\nonumber\\
  &&(x_o[k]-x_{obst}[k]) \sin \frac{2 \pi m}{M_{o}}\nonumber\\ 
  &&{+}\:(y_o[k]-y_{obst}[k]) \cos \frac{2 \pi m}{M_{o}} >
    R_{obst}.
\end{eqnarray}

To express this avoidance constraint in a MILP problem formulation, it
must be converted to an equivalent set of linear inequality
constraints. We do so by introducing auxiliary binary variable
$b_m[k] \in \{ 0,1 \}$ and the following $M_o$ inequality constraints,
\begin{eqnarray}
\label{avoidconstraints1}
  &&(x_o[k]-x_{obst}[k]) \sin \frac{2 \pi m}{M_{o}}\nonumber\\
  &&{+}\:(y_o[k]-y_{obst}[k]) \cos \frac{2 \pi m}{M_{o}}
  > R_{obst} - H b_m[k]\nonumber\\
  &&\forall m \in \{ 1,\ldots,M_{o} \}, 
\end{eqnarray}
where $H$ is a large positive number taken to be larger than the
maximum dimension of the vehicle's operating environment plus the
radius of the obstacle.  If $b_m[k] = 1$, the right hand side of the
inequality is a large, negative number that is always less than the
left hand side.  In this case, the inequality is inactive because it
is trivially satisfied.  If $b_m[k]= 0$, the inequality is said to be
active because it reduces to an inequality from the existence
condition above.
For obstacle avoidance, at least
one of the constraints in equation~(\ref{avoidconstraints1}) must be
active. To enforce this, we introduce the following inequality
constraint into the problem formulation,
\begin{eqnarray}
  \sum_{m=1}^{M_{o}} b_m[k] \leq M_{o}-1.
  \label{avoidconstraints2}
\end{eqnarray}

Therefore, to enforce obstacle avoidance at time $t_o[k]$, the set of
binary variables $\{b_m[k]\}_{m=1}^{M_o}$ and the constraints given by
equations~(\ref{avoidconstraints1}) and~(\ref{avoidconstraints2}) are
added to the MILP problem formulation.

Consider the example problem from Section~\ref{vehicle_dynamics} with
obstacles. In this problem, we want to transfer the vehicle from start
state $\mathbf{x}_s$ to finish state $\mathbf{x}_f$ in time $t_f$
using minimal control effort while avoiding obstacles.
To enforce obstacle avoidance at each
time in the set $\{t_o[k]\}_{k=1}^{N_o}$, we augment the MILP
formulation in Section~\ref{vehicle_dynamics}
with the set of binary variables $\{b_m[k]\}_{m=1}^{M_o}$,
constraints~(\ref{avoidconstraints1}), and
constraint~(\ref{avoidconstraints2}) for all $k$ in the set
$\{1,\ldots,N_o\}$.

\begin{figure}
\centering
\includegraphics[width=3.5in]{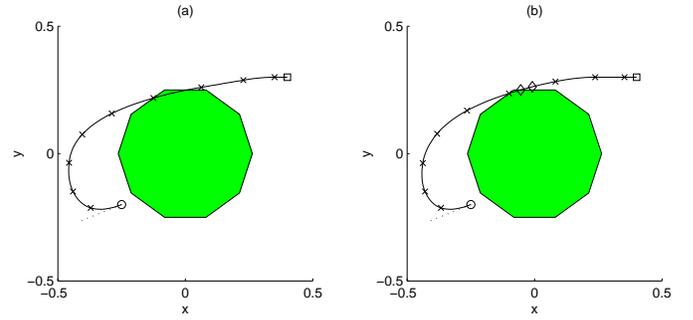}
\caption{Plots of the minimum control effort obstacle 
avoidance example.  The shaded region is the obstacle to be avoided
and the $\times$'s along the trajectory denote the avoidance points
$(x_o[k],y_o[k])$.  Figure~(a) is the original solution.  Figure~(b)
is the solution after two steps of the iterative obstacle avoidance
algorithm. The $\Diamond$'s are the avoidance
points added to the MILP formulation by the iterative algorithm.  The
values for the parameters are $N_u = 10$, $M_u = 20$, $T_u[k] =
0.3$ for all $k$, $M_{o} = 10$, $N_{o} = 10$, $t_o[k] = k
T$, $(x_s,y_s,\dot{x}_s,\dot{y}_s) = (-0.25,
-0.2, -0.5, -0.2)$, and $(x_f,y_f,\dot{x}_f,\dot{y}_f) = (0.4, 0.3,
0.0, 0.0)$.
} 
  \label{mineffortobst}
\end{figure}

Distributing the avoidance times uniformly (uniform gridding), as
in~\cite{Richards02c,Earl02}, results in a trajectory that avoids
obstacles at each discrete time in the set,  but the trajectory can 
collide with obstacles between avoidance times. This is shown for
an example instance in Figure~\ref{mineffortobst}(a). In this example,
the trajectory intersects the obstacle between the sixth and
seventh avoidance time steps.  A simple method to eliminate this
behavior is to represent the obstacle with a polygon that is larger
than the obstacle. Then distribute obstacle avoidance times uniformly
such that the sampling time is small enough to guarantee avoidance.
In general, this is not a desirable approach because it results in
large MILPs that require significant computational effort to solve.

A better approach is to select the avoidance times intelligently.  
In~\cite{Earl04b}, we
have developed an iterative MILP algorithm that does this.
We summarize this algorithm here.
First, pick an initial set of times $\{t_o[k]\}_{k=1}^{N_o}$ at which obstacle
avoidance will be enforced.  Then, formulate and solve the MILP as
described above representing the obstacles with polygons
slightly larger than the obstacles.  Next, check the resulting
trajectory for collisions with any of the obstacles (not the polygons
which represent them in the MILP).  If there are no collisions,
terminate the algorithm.  If there is a collision, compute the time
intervals for which collisions occur denoting the time interval for
collision $i$ by $(t_a^{(i)},t_b^{(i)})$.  For each interval $i$, pick
a time within the interval, such as $(t_a^{(i)}+t_b^{(i)})/2$.  At
each of these times add obstacle avoidance constraints to the MILP
formulation.  Then, solve the augmented MILP repeating the procedure
above (first checking if the resulting trajectory intersects any
obstacles, etc.) until a collision free trajectory is found.
Figure~\ref{mineffortobst}(b) shows the effectiveness of this
algorithm after two iterations. 

\section{RoboFlag Problems}
\label{sec:dd}
\begin{figure}
\centering
\includegraphics[width=3.0in]{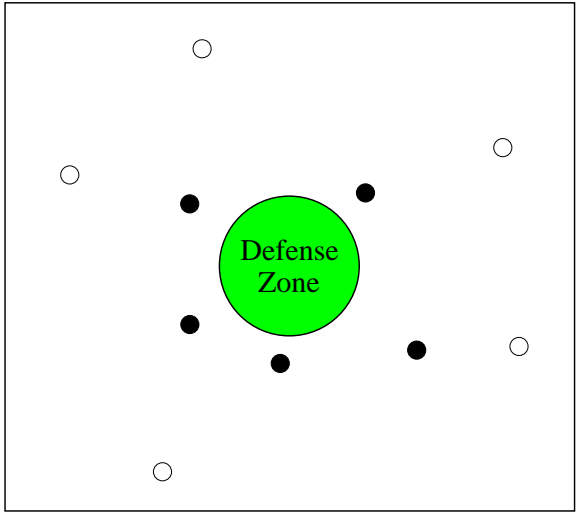}
\caption{The RoboFlag Drill used to
motivate the methods presented is this paper.
The drill takes place on a playing field with a Defense Zone at its
center. The objective is to design a cooperative control strategy for
the team of defending vehicles (black) that minimizes the number of
attacking vehicles (white) that enter the Defense Zone.}
\label{drill}
\end{figure}
To motivate our multi-vehicle methods, we apply them to simplified
versions of the RoboFlag
game~\cite{d'andrea03a,d'andrea03b,campbell03}, which we call RoboFlag
Drills because they serve as practice for the real game.  The drills
involve two teams of robots, the defenders and the attackers, on a
playing field with a circular region of radius $R_{dz}$ at its center
called the Defense Zone, as shown in Figure~\ref{drill}.   The
attackers' objective is to fill the Defense Zone with as many
attackers as possible.  The defenders' objective is to deny as many
attackers from entering the Defense Zone as possible without entering
the zone themselves.  We consider Defensive Drill problems in which
each attacker has a fixed intelligence governed by a state machine.
The goal is to design a team control strategy for the defenders that
maximizes the number of attackers denied from the Defense Zone.  In
this section, we use MILP methods to generate such strategies.
We consider two versions of the Defensive Drill each with a different
set of laws governing attacker intelligence.

To start, we consider one-on-one Defensive Drill problems.  This is the
simplest case and involves one defender and one attacker.  Although
this case is not particularly interesting, we start with it for
notational clarity.  Next, we generalize to the case involving $N_D$
defenders and $N_A$ attackers, which is a straightforward extension.

\subsection{Defensive Drill 1: one-on-one case}
\label{sec:oneonone}
The defender is governed by the discrete time dynamical system from 
Section~\ref{vehicle_dynamics}
\begin{eqnarray}
  &&\mathbf{x}_u[k+1] = \mathbf{A}[k] \mathbf{x}_u[k] 
  + \mathbf{B}[k]
  \mathbf{u}[k]\nonumber\\
  &&\mathbf{x}_u[0] = \mathbf{x}_s\nonumber\\
  &&u_{x}[k] \sin \frac{2 \pi m}{M_u} + 
  u_{y}[k] \cos \frac{2 \pi m}{M_u} \leq 
  \cos \frac{\pi}{M_u}\nonumber\\ 
  &&\forall m \in \{ 1,\ldots,M_u \}\nonumber \\
  &&\forall k \in \{ 1,\ldots,N_u \}.
  \label{defdyn2}
\end{eqnarray}

The attacker has two discrete modes: attack
and inactive.  When in attack mode, it moves toward
the Defense Zone at constant velocity along a straight line path.  The
attacker, initially in attack mode at the beginning of play,
transitions to inactive mode if the defender intercepts it or if it
enters the Defense Zone.   Once inactive, the attacker does not move
and remains inactive for the remainder of play.
These dynamics are captured by the
following discrete time equations and state machine 
\begin{eqnarray}
  &&p[k+1] = p[k] + v_p T_a[k] a[k]\nonumber\\
  &&q[k+1] = q[k] + v_q T_a[k] a[k]
  \label{attdyn1}
\end{eqnarray}
\begin{eqnarray}
  &&a[k+1] = \left\{
  \begin{array}{ll}
    1 & \mbox{if ($a[k]=1$)} \\
      & \mbox{and (not in Defense Zone)} \\ 
      & \mbox{and (not intercepted) } \\
    0 & \mbox{if ($a[k]=0$)}\\
      & \mbox{or (in Defense Zone)} \\
      & \mbox{or (intercepted)}
  \end{array} \right. \\
  &&\forall k \in \{ 1,\ldots,N_a \} \nonumber
\end{eqnarray}
\begin{figure}
\centering
\includegraphics[width=2.5in]{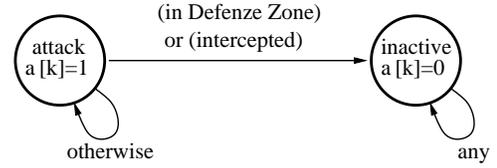}
\caption{The two state (attack and inactive) attacker state machine.
The attacker starts in the attack state. It transitions to the
inactive state, and remains in this state, if it enters the Defense
Zone or if it is intercepted by the defender.} 
\label{sm1}
\end{figure}
with initial conditions
\begin{eqnarray}
  \mbox{$p[0] = p_s$, $q[0] = q_s$, and $a[0] = 1$},  
  \label{attdyn2}
\end{eqnarray}
where $N_a$ is the number of samples, $k \in \{ 1,\ldots, N_a\}$,
$T_a[k]>0$ is the time between samples $k$ and $k+1$, $(p[k],q[k])$ is
the attacker's position at time $t_a[k] = \sum_{i=0}^{k-1} T_a[i]$,
$(v_p,v_q)$ is its constant velocity vector, and $a[k] \in \{0,1 \}$
is a discrete state indicating the attacker's mode.  The attacker is
in attack mode when $a[k] = 1$ and inactive mode when $a[k] = 0$.
The attacker state machine is shown in Figure~\ref{sm1}. Here we use the
subscript $a$ to denote the time discretization for the attacker's
dynamics.
\\

\noindent {\bf \emph{Defense Zone}}

Because the defender wants to keep the attacker from entering the
Defense Zone, a binary variable indicating whether or not the
attacker is inside the Defense Zone is introduced.  This variable
is used to define the attacker state machine precisely.  We
denote the binary variable with $\gamma[k] \in \{ 0,1 \}$.  When the
attacker is in the Defense Zone at step $k$, $\gamma[k]=1$. When
the attacker is outside the Defense Zone at step $k$, $\gamma[k]=0$. 

Similar to the approach used to define obstacles, the Defense Zone is
approximated using a polygon $\mathcal{G}$ defined by a set of
$M_{dz}$ inequalities
\setlength{\arraycolsep}{0.0em}
\begin{eqnarray}
  \mathcal{G}&{}:={}& \{ \mbox { }(\bar{x},\bar{y}):\nonumber\\ 
  &&\bar{x} \sin \frac{2 \pi m}{M_{dz}} + 
  \bar{y} \cos \frac{2 \pi m}{M_{dz}} \leq 
  R_{dz}\nonumber\\
  &&\forall m \in \{1,\ldots,M_{dz}\} \mbox{ } \}.
\end{eqnarray}
\setlength{\arraycolsep}{5pt}%
The association between $\gamma[k]$ and $\mathcal{G}$ is
\begin{equation}
  (\gamma[k] = 1) \iff (p[k],q[k]) \in \mathcal{G}.
  \label{gamma1}
\end{equation}
If the defender keeps the binary variable $\gamma[k]$ equal
to $0$ for
all $k \in \{ 1,\ldots, N_a \}$, it has successfully denied the
attacker from the region $\mathcal{G}$ and thus from the Defense Zone.
However, in order to use the binary variable $\gamma[k]$ in the problem
formulation, we must enforce the logical constraint given by
equation~(\ref{gamma1}).  To enforce this constraint in MILP, we
convert it into an equivalent set of inequality constraints.

We introduce the binary variable $g_m[k]\in \{ 0,1 \}$
to indicate whether or not the $m$th constraint of $\mathcal{G}$ is
satisfied by the attacker with position $(p[k],q[k])$.  This
association is made by introducing the logical constraint
\begin{eqnarray}
  \label{gequiv}
  &&\left( g_m[k] = 1 \right) \iff \nonumber\\
  &&\left(
  p[k] \sin \frac{2 \pi m}{M_{dz}} + q[k] \cos \frac{2 \pi m}{M_{dz}} \leq 
  R_{dz} 
  \right).
\end{eqnarray}
As shown in Appendix~\ref{logicconvert}, 
it is equivalent to the following set of inequalities
\setlength{\arraycolsep}{0.0em}
\begin{eqnarray}
  &&p[k] \sin \frac{2 \pi m}{M_{dz}}
  +q[k] \cos \frac{2 \pi m}{M_{dz}}\nonumber\\
  &&\;\;\;\;\leq R_{dz} + H (1-g_m[k])\nonumber\\
  &&p[k] \sin \frac{2 \pi m}{M_{dz}}
  +q[k] \cos \frac{2 \pi m}{M_{dz}}\nonumber\\ 
  &&\;\;\;\;\geq R_{dz} + \epsilon - (H+\epsilon) g_m[k],
  \label{gammacon1}
\end{eqnarray}
\setlength{\arraycolsep}{5pt}%
where $\epsilon$ is a small positive number and $H$ is a large positive
number such that the left hand sides of the inequalities are bounded
from above by $H$ and from below by $-H$.

Using binary variable $g_m[k]$, we can write
equation~(\ref{gamma1}) as
\begin{eqnarray}
  &&(\gamma[k] = 1) \iff\nonumber\\
  &&(g_m[k] = 1\;\;\;\;
  \forall m \in \{1,\ldots,M_{dz} \}),
  \label{gamma2}
\end{eqnarray}
which is equivalent to the inequality constraints
\begin{eqnarray}
   &&g_m[k] - \gamma[k] \geq 0\;\;\;\; 
   \forall m \in \{ 1,\ldots,M_{dz} \}\nonumber\\ 
   &&\sum_{i=1}^{M_{dz}}(1-g_i[k]) + \gamma[k] \geq 1,
   \label{gammacon2}
\end{eqnarray}
as shown in Appendix~\ref{logicconvert}. 

The logical constraint given by equation~(\ref{gamma1}) is equivalent
to the inequality constraints given by equations~(\ref{gammacon1})
and~(\ref{gammacon2}).  Therefore, we can include the indicator
variable $\gamma[k]$ in the MILP formulation by also including the binary
variables $\{ g_m[k] \}_{m=1}^{M_a}$ and constraints~(\ref{gammacon1})
and~(\ref{gammacon2}).
\\

\noindent {\bf \emph{Intercept Region}}

To define what it means for a defender to intercept an attacker,
we introduce an intercept region $\mathcal{I}[k]$ rigidly attached
to the defending robot.  The intercept region is a polygon defined by
a set of $M_I$ inequalities.  If an attacker is in this polygon, it is
considered intercepted.  For the defender with coordinates
$(x_a[k],y_a[k])$, the intercept region is given by
\setlength{\arraycolsep}{0.0em}
\begin{eqnarray}
  \mathcal{I}[k]&{}:={}& \{ \mbox { }(\bar{x},\bar{y}):\nonumber\\ 
  &&(\bar{x} - x_a[k]) \sin \frac{2 \pi m}{M_I} + 
  (\bar{y} - y_a[k]) \cos \frac{2 \pi m}{M_I} \leq R_I\nonumber\\
  &&\forall m \in \{1,\ldots,M_I\} \mbox{ } \},  
\end{eqnarray}
\setlength{\arraycolsep}{5pt}%
where $x_a[k] = x(t_a[k])$ and $y_a[k] = y(t_a[k])$ are 
calculated using equation~(\ref{inbetween}),
and $R_I$ is the inscribed radius of the polygon.  

The binary variable $\delta[k] \in \{0,1\}$ is introduced to indicate
whether or not the attacker is inside the defender's intercept region. The
association is made with the following logical constraint
\begin{eqnarray}
  \label{intercept1}
  \delta[k] = 1 \iff (p[k],q[k]) \in \mathcal{I}[k].
\end{eqnarray}
Using techniques similar to those used for $\gamma[k]$, we convert this
constraint into an equivalent set of inequality constraints as
follows:
For each of the
inequalities defining the intercept region, we associate a binary variable
$d_m[k] \in \{ 0,1 \}$ by introducing the logical constraint
\begin{eqnarray}
  &&\left( d_m[k] = 1 \right) \iff \nonumber\\ 
    &&\left((p[k] - x_a[k]) \sin \frac{2 \pi m}{M_I}\right.\nonumber \\ 
    &&{+}\:\left.(q[k] - y_a[k]) \cos \frac{2 \pi m}{M_I} \leq
    R_{I}\right),
\end{eqnarray}
where $(p[k],q[k])$ are the coordinates of the attacking robot.
If $d_m[k] = 1$, the coordinates of the attacking robot satisfy the
$m$th inequality defining the intercept region. 
If $d_m[k] = 0$, the $m$th inequality is not satisfied.  Similar to
equation~(\ref{gequiv}), we can express this logic constraint as the
set of inequalities
\setlength{\arraycolsep}{0.0em}
\begin{eqnarray}
  &&(p[k] - x_a[k]) \sin \frac{2 \pi m}{M_I} 
  +(q[k] - y_a[k]) \cos \frac{2 \pi m}{M_I}\nonumber\\ 
  &&\;\;\;\;\leq R_I +  H(1-d_m[k])\nonumber\\
  &&(p[k] - x_a[k]) \sin \frac{2 \pi m}{M_I} 
  +(q[k] - y_a[k]) \cos \frac{2 \pi m}{M_I}\nonumber\\ 
  &&\;\;\;\;\geq R_I + \epsilon - (H+\epsilon) d_m[k].
  \label{deltacon1}
\end{eqnarray}
\setlength{\arraycolsep}{5pt}%
Using the binary variables $d_m[k]$ we can write equation~(\ref{intercept1}) as 
\begin{eqnarray}
  \label{intercept2}
  \delta[k] = 1 \iff 
  (d_m[k] = 1 & \forall m \in \{1,\ldots,M_I \}).
\end{eqnarray}
Considering both directions of this equivalence, as done for
$\gamma[k]$ above, we find that the statement is equivalent to the
following set of inequality constraints
\begin{eqnarray}
  &&d_m[k] - \delta[k] \geq 0\;\;\;\;
  \forall m \in \{ 1,\ldots,M_I\}\nonumber\\
  &&\sum_{i=1}^{M_I}(1-d_i[k]) + \delta[k] \geq 1.  
  \label{deltacon2}
\end{eqnarray}
We can use $\delta[k]$ in our problem formulation if we include the
binary variables $\{d_m[k]\}_{m=1}^{M_a}$ and the inequalities
constraints given by equations~(\ref{deltacon1})
and~(\ref{deltacon2}).
\\
\\

\noindent {\bf \emph{State machine and objective function} }

With the indicator variables $\gamma[k]$ and $\delta[k]$ defined, we
revisit the attacker state machine and define it more precisely with
the following state equation
\begin{figure}
\centering
\includegraphics[width=2.5in]{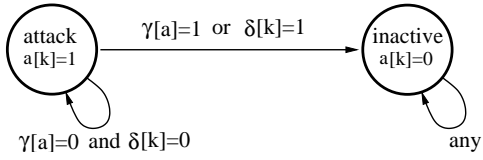}
\caption{The two state attacker state machine, as in Figure~\ref{sm1},
using binary variables $\delta[k]$ and $\gamma[k]$.} 
\label{sm2}
\end{figure}
\begin{eqnarray}
a[k+1] = \left\{
\begin{array}{ll}
1 & \mbox{if $a[k]=1$ and $\gamma[k]=0$}\\
  & \mbox{and $\delta[k]=0$} \\
0 & \mbox{if $a[k]=0$ or $\gamma[k]=1$}\\
  & \mbox{or $\delta[k]=1$},
\end{array} \right.
\label{asgov2}
\end{eqnarray}
which is shown in Figure~\ref{sm2}.  The state equations
can be written as the logical expression
\begin{eqnarray}
  &&\mbox{$(a[k+1] = 1) \iff$}\nonumber\\
  &&\mbox{$(a[k]=1$ and $\gamma[k]=0$ and $\delta[k]=0)$},
  \label{asgovlogic}
\end{eqnarray}
which is equivalent to the set of inequality constraints
\begin{eqnarray}
  &&a[k+1] + \delta[k] \leq 1\nonumber\\
  &&a[k+1] - a[k] \leq 0\nonumber\\
  &&a[k+1] + \gamma[k] \leq 1\nonumber\\
  &&a[k] - \delta[k] - \gamma[k] - a[k+1] \leq 0,
  \label{attdyn3}
\end{eqnarray}
as shown in Appendix~\ref{logicconvert}. 

We have defined the dynamics of the defenders and the attackers, and
we have converted these dynamics to MILP form.  The final step in
formulating Defensive Drill 1 is to introduce an objective function
that captures the defender's desire to deny the attacker from entering
the Defense Zone. This objective is achieved by minimizing the
binary variable $\gamma[k]$ over the duration of the drill, which is
captured by minimizing the function  
\begin{eqnarray}
  J = \sum_{k=1}^{N_a} \gamma[k].
\end{eqnarray}
We set the duration of the drill such that
\begin{eqnarray}
  t_a[N_a] \geq \sqrt{ \frac{p_s^2 + q_s^2}{v_p^2 + v_q^2} }.
\end{eqnarray}
This allows the attacker enough time to reach the Defense Zone if it is
not intercepted.  

In addition to intercepting the attacker, we would
like to conserve fuel as a secondary objective.  One way to achieve
this is to add a small penalty on the control effort to the objective
function as follows:
\begin{eqnarray}
  J = \sum_{k=1}^{N_a} \gamma[k] 
  + \epsilon \sum_{k =0}^{N_u-1} 
  \left( |u_{x}[k]| + |u_{y}[k]| \right),
  \label{objective3}
\end{eqnarray}
where $\epsilon$ is taken to be a small positive number because we
want the minimization of the binary variable $\gamma[k]$ to dominate.
We use the procedure outlined in Section~\ref{vehicle_dynamics} to
write equation~(\ref{objective3}) in MILP form.
\\

\noindent {\bf \emph{Summary and example}}

We have formulated Defensive Drill 1 as the
following optimization problem:  minimize~(\ref{objective3}) 
subject to defender dynamics~(\ref{defdyn2});
attacker dynamics~(\ref{attdyn1}),~(\ref{attdyn2}),
and~(\ref{attdyn3}) for all $k \in \{ 1,\ldots N_a\}$;
the constraints for the $\gamma[k]$ indicator
variable~(\ref{gammacon1}) and~(\ref{gammacon2})
for all $m \in \{1,\ldots,M_{dz}\}$ and for all $k \in \{ 1,\ldots,
N_a\}$;
the constraints for the $\delta[k]$ indicator
variable~(\ref{deltacon1}), and~(\ref{deltacon2})
for all $m \in \{1,\ldots,M_{I}\}$ and for all $k \in \{1,\ldots, N_a \}$;
and the avoidance constraints for the 
Defense Zone~(\ref{avoidconstraints1}) and~(\ref{avoidconstraints2}) 
for all $m \in \{ 1,\ldots,M_{o} \}$ and for all $k \in \{
1,\ldots,N_o\}$.

\begin{figure}
\centering
\includegraphics[width=3.0in]{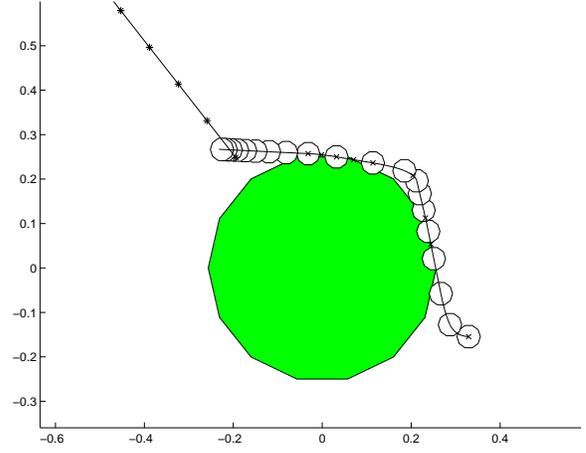}
\caption{The defender denies the attacker from entering the 
Defense Zone.  The $\ast$'s along the attacker's trajectory
denote the attacker's position for each time $t_a[k]$.  The polygons 
along the defender's trajectory denote the intercept region
$\mathcal{I}[k]$ for each time $t_a[k]$.  The $\times$'s denote
obstacle avoidance points. 
}  
  \label{1d1a}
\end{figure}

The solution to an instance of this problem is shown in
Figure~\ref{1d1a}.  For this
instance, the defender denies the attacker from the Defense Zone
(shaded polygon) by intercepting it.  Each asterisk along the attacker's
trajectory denotes the position of the attacker at sample time 
$t_a[k]$.  The
white polygons along the defender's trajectory denotes the 
intercept region $\mathcal{I}[k]$ at time $t_a[k]$.  

\subsection{Defensive Drill 2: one-on-one case}
\label{sec:dd2oneonone}
\begin{figure}
\centering
\includegraphics[width=3.0in]{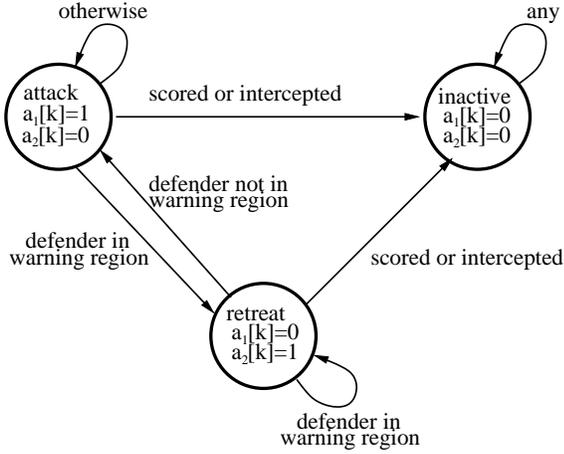}
\caption{The three state (attack, inactive, and retreat) attacker
state machine.  The attacker starts in the attack state.}
\label{sm3}
\end{figure}
The dynamics of the defender are the same as the defender dynamics in
Section~\ref{sec:oneonone}.  The dynamics of the attacker are the same
as the attacker dynamics in Defensive Drill 1, but with an additional
state called retreat.  If a defender is too close to the attacker, the
attacker enters the retreat state and reverses direction.  While
retreating, if the defender is far enough away, the attacker returns
to attack mode.

These dynamics are captured by the following discrete 
time equations and state machine:
\begin{eqnarray}
  &&p[k+1] = p[k] + v_p T_a[k] a_1[k]\nonumber\\
  &&\;\;\;\;-\: v_p T_a[k] a_2[k]\nonumber\\
  &&q[k+1] = q[k] + v_q Ta[k] a_1[k]\nonumber\\ 
  &&\;\;\;\;-\: v_q T_a[k] a_2[k] 
  \label{dd2attdyn1}
\end{eqnarray}
\begin{eqnarray}
  &&a_1[k+1] = \left\{
  \begin{array}{ll}
    1 &  
        \mbox{if } [ \mbox{($a_1[k]=1$ and $a_2[k]=0$)} \\ 
      & \mbox{or ($a_1[k]=0$ and $a_2[k]=1$)} ] \\ 
      & \mbox{and (not scored)} \\
      & \mbox{and (not intercepted)} \\ 
      & \mbox{and (not in warning region) } \\
    0 & \mbox{otherwise}
  \end{array} \right.\\
  &&a_2[k+1] = \left\{
  \begin{array}{ll}
    1 &  
        \mbox{if } [ \mbox{($a_1[k]=0$ and $a_2[k]=1$) or }  \\
      & \mbox{($a_1[k]=1$ and $a_2[k]=0$)} ] \\
      & \mbox{and (not scored)} \\
      & \mbox{and (not intercepted)} \\
      & \mbox{and (in warning region) } \\
    0 & \mbox{otherwise}
  \end{array} \right.
\end{eqnarray}
for all $k \in \{ 1,\ldots,N_a \}$ and subject to the constraint
\begin{eqnarray}
\label{dd2ascon1}
&a_1[k] + a_2[k] \leq 1
\end{eqnarray}
and the initial conditions
\begin{eqnarray}
&\mbox{$p[0] = p_s$, $q[0] = q_s$, $a_1[0] = 1$, $a_2[0] = 0$}. 
  \label{dd2attic}
\end{eqnarray}
The state machine is shown in Figure~\ref{sm3}.  

The attacker needs two binary state variables because it has three
discrete modes of operation: attack, retreat, and inactive. These
modes are represented by $(a_1[k],a_2[k]) = (1,0)$, $(a_1[k],a_2[k]) =
(0,1)$, and $(a_1[k],a_2[k]) = (0,0)$, respectively.  Because the
state $(a_1[k],a_2[k]) = (1,1)$ is not needed, we impose the
inequality constraint given by equation~(\ref{dd2ascon1}).

To determine if the defender is too close to the attacker, a warning
region is introduced.  The warning region $\mathcal{W}[k]$ is a
polygon defined by a set of inequalities similar to the intercept
region $\mathcal{I}[k]$. We introduce binary auxiliary variables 
$w_m[k]$ and $\omega[k]$, and we introduce inequality constraints
similar to equations~(\ref{deltacon1}) and~(\ref{deltacon2}). The result
is the following association for indicator variable $\omega[k]$:
If $\omega[k] = 1$, the defender is in the attacker's
warning region at step $k$, otherwise, $\omega[k] = 0$.
The attacker state machine can be written as the set of inequality
constraints
\begin{eqnarray}
  &&a_1[k+1] - a_1[k] + a_2[k] + \gamma[k] + \delta[k] + \omega[k]
    \geq 0\nonumber\\
  &&a_1[k+1] + a_1[k] - a_2[k] + \gamma[k] + \delta[k] + 
     \omega[k] \geq 0\nonumber\\ 
  &&a_2[k+1] + a_1[k] - a_2[k] + \gamma[k] + \delta[k] - 
    \omega[k] \geq -1\nonumber\\
  &&a_2[k+1] - a_1[k] + a_2[k] + \gamma[k] + \delta[k] - 
    \omega[k] \geq -1\nonumber\\ 
  &&a_1[k+1] + a_1[k] + a_2[k] \leq 2\nonumber\\ 
  &&a_1[k+1] - a_1[k] - a_2[k] \leq 0\nonumber\\
  &&a_1[k+1] + \gamma[k] \leq 1\nonumber\\
  &&a_1[k+1] + \delta[k] \leq 1\nonumber\\
  &&a_1[k+1] + \omega[k] \leq 1\nonumber\\ 
  &&a_2[k+1] - a_1[k] - a_2[k] \leq 0\nonumber\\ 
  &&a_2[k+1] + a_1[k] + a_2[k] \leq 2\nonumber\\
  &&a_2[k+1] + \gamma[k] \leq 1\nonumber\\
  &&a_2[k+1] + \delta[k] \leq 1\nonumber\\  
  &&a_2[k+1] - \omega[k] \leq 0.  
  \label{dd2asineq}
\end{eqnarray}

\begin{figure}
\centering
\includegraphics[width=3.5in]{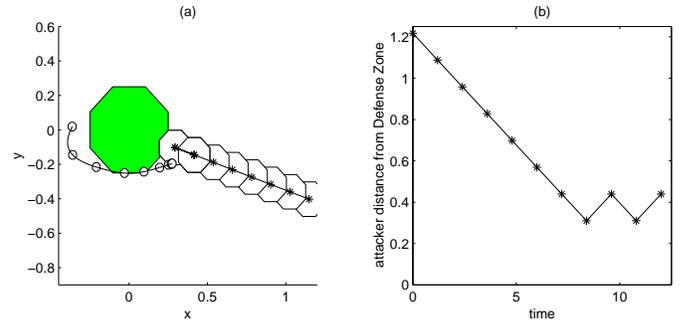}
\caption{The solution to an instance of Defensive Drill 2. 
Figure~(a) shows the playing field.  Each polygon along the attacker's
trajectory is a warning region.  
Figure~(b) shows the attacker's distance
from the center of the Defense Zone versus time.
The parameters are
$M_o=8$, $M_u=4$, $M_I=8$, $M_w=8$, $M_{dz}=8$, $N_u=4$, $N_o=4$, $N_a=10$,
$\mathbf{x}_0 = (-.36,.02,-.12,-.22)$, and
$(p_0,q_0,v_a)=(1.15,-.4,-.11)$.}
  \label{dd2_1d1a_1}
\end{figure}

Similar to Defense Drill 1, Defense Drill 2 can be posed as a MILP.
The solution of an example instance is shown in Figure~\ref{dd2_1d1a_1}.

\subsection{$N_D$-on-$N_A$ case}
\label{sec:dd1general}
To generalize the problem formulation to $N_D$ defenders and $N_A$
attackers, we need to add more variables.  Defender $i$, where
$i\in\{1,\ldots,N_D\}$, has state $\mathbf{x}_{ui}[k]$, control input
$\mathbf{u}_{i}[k]$, and slack variables $z_{xi}[k]$ and $z_{yi}[k]$.
Attacker $j$, where $j\in\{1,\ldots,N_A\}$, has state
$(p_j[k],q_j[k],a_j[k])$ and constant velocity vector
$(v_{pj},v_{qj})$.  The binary variable $\gamma_j[k]$ equals $1$ if
and only if attacker $j$ is in polygon $\mathcal{G}$ at step $k$.  The
binary variable $\delta_{ij}[k]$ equals $1$ if and only if attacker $j$ is in
defender $i$'s intercept region $\mathcal{I}_i[k]$ at step $k$. The
binary variables $g_{mj}[k]$, $d_{mij}[k]$, and $b_{mi}[k]$ follow a
similar trend.  For Defensive Drill 2, we also need $\omega_{ij}[k] =
1$ if and only if defender $i$ is in attacker $j$'s warning region
$\mathcal{W}_j[k]$ at step $k$.  And similarly for the binary variable
$w_{mij}[k]$. 

With the variables for the general case defined, inequality constraints
are added to the formulation in a similar way as those derived for
the one-on-one case.  For example, the constraints identifying
$\gamma_j[k] = 1$ with $(p_j[k],q_j[k])\in\mathcal{G}$ are
\begin{eqnarray}
   &&g_{mj}[k] - \gamma_j[k] \geq 0\;\;\;\; 
   \forall m \in \{ 1,\ldots,M_{dz} \}\nonumber\\ 
   &&\sum_{l=1}^{M_{dz}}(1-g_{lj}[k]) 
     + \gamma_j[k] \geq 1,\nonumber
\end{eqnarray}
for all $k \in \{ 1,\ldots,N_a\}$ and for all $j \in \{1,\ldots,N_A\}$.

And finally, the objective function is given by
\begin{eqnarray}
\label{eqn:objective}
  J = \sum_{j=1}^{N_A} \sum_{k=1}^{N_a} \gamma_j[k] 
  + \epsilon \sum_{i=1}^{N_D} \sum_{k =0}^{N_u-1} 
  \left( |u_{xi}[k]| + |u_{yi}[k]| \right).
\end{eqnarray}

\begin{figure}
\centering
\includegraphics[width=3.5in]{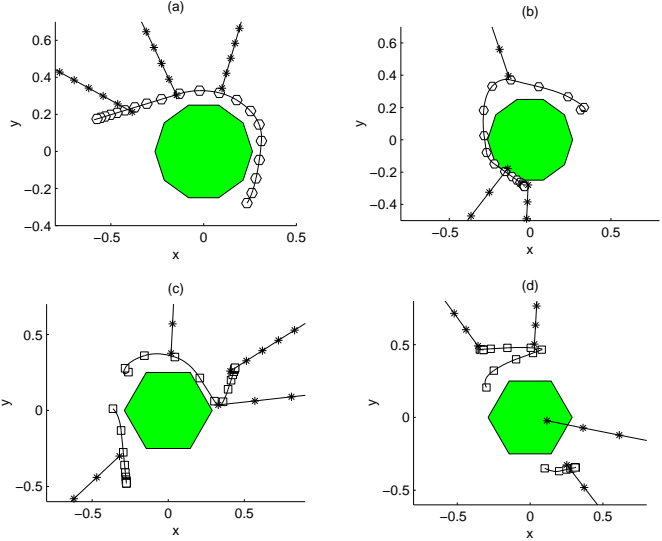}
\caption{The solution to four instances of
Defensive Drill 1.  
For Figures~(a) and (b), $N_A=3$ and $N_D=1$.
For Figures~(c) and (d), $N_A=4$ and $N_D=2$.  
In Figures (a) and (c), the defenders deny all attackers from the
Defense Zone.  
In Figures~(b) and (d), an 
attacker enters the Defense Zone.
}
  \label{1d3a}
\end{figure}

\begin{figure}
\centering
\includegraphics[width=3.5in]{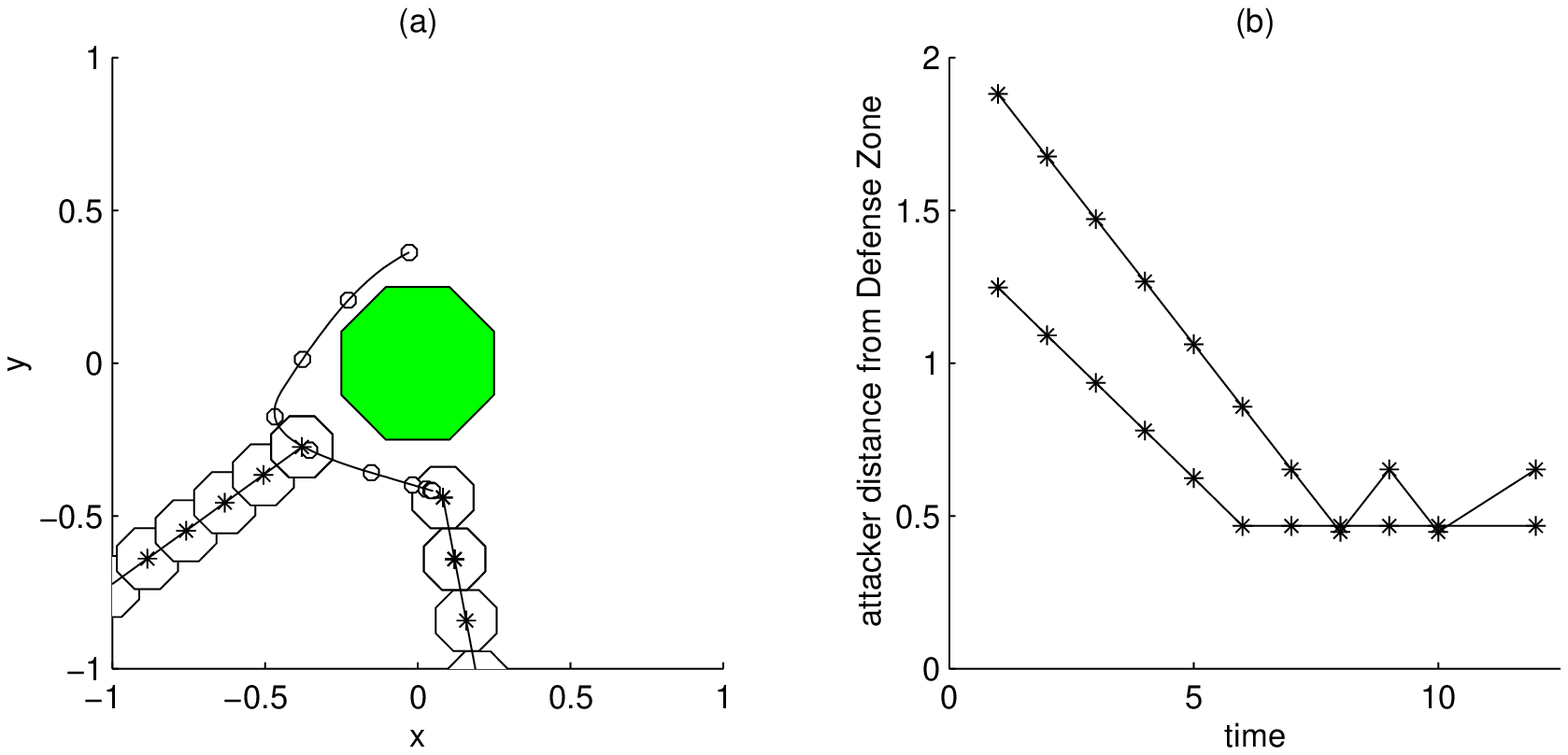}
\includegraphics[width=3.5in]{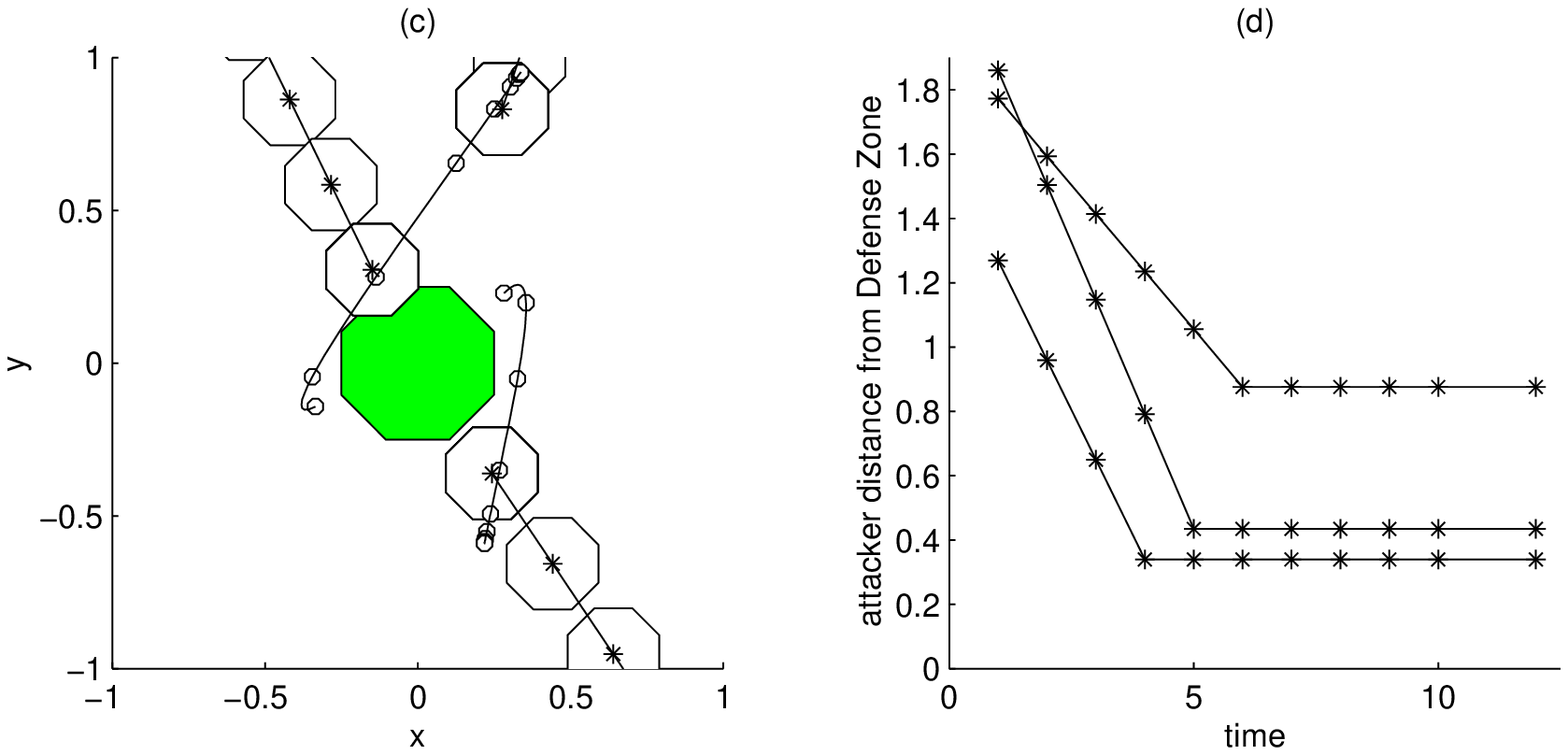}
\caption{The solution to two instances of 
Defensive Drill 2.  
Figures~(a) and (c) show the playing field.  
Figures~(b) and (d) show each attacker's distance
from the center of the Defense Zone versus time.
}
\label{dd2_1d2a_1}
\end{figure}

Results for example instances of Defensive Drill 1 are shown 
in Figure~\ref{1d3a}.  
Results for Defensive Drill 2 are shown in Figure~\ref{dd2_1d2a_1}.
\label{sec:dd2general}

\section{Average case complexity}
\label{sec:ctime}
In this section, we explore the average case complexity of Defensive
Drill 1 by solving randomly generated instances.  Each instance is
generated by randomly picking parameters from a uniform distribution
over the intervals defined below.  Each MILP is solved using
AMPL~\cite{fourer93} and CPLEX~\cite{ilog00} on a PC with Intel PIII
550MHz processor, 1024KB cache, 3.8GB RAM, and Red Hat Linux.
For all instances solved, processor speed was the limiting factor, not
memory.

To generate the initial condition $(p_s,q_s)$ and the constant
velocity vector $(v_p,v_q)$ for each attacker we pick $r_a$,
$\theta_a$, and $v_a$ randomly from a uniform distribution over the
intervals $[r_a^{\min},r_a^{\max}]$, $[0,2 \pi)$, and
$[v_a^{\min},v_a^{\max}]$, respectively.  The parameters are then
given by
\begin{eqnarray}
{}&{}&p_s = r_a \cos \theta_a,\mbox{ }q_s = r_a \sin \theta_a\nonumber\\
{}&{}&v_p =  v_a p_s/ \sqrt{p_s^2 + q_s^2},\mbox{ }
v_q =  v_a q_s/ \sqrt{p_s^2 + q_s^2}. 
\end{eqnarray}

To generate the initial condition $(x_s,y_s,\dot{x}_s,\dot{y}_s)$ for
each defender, we pick $r_d$, $\theta_d$, $v_d$, and $\theta_v$
randomly from a uniform distribution over the intervals
$[r_d^{\min},r_d^{\max}]$, $[0,2 \pi)$, $[v_d^{\min},v_d^{\max}]$, and
$[0,2 \pi)$, respectively. The defender's initial condition is then
given by
\begin{eqnarray}
&&x_s = r_d \cos \theta_d,\mbox{ }y_s = r_d \sin \theta_d\nonumber\\
&&\dot{x}_s = v_d \cos \theta_v,\mbox{ }
\dot{y}_s = v_d \sin \theta_v.
\end{eqnarray}

We take the playing field to be circular with radius $R_f = 15$, and
we set the radius of the Defense Zone to be $R_{dz} = 2$.
The intervals used to generate the instances are 
$r_d \in [\sqrt{2}R_{dz},2\sqrt{2}R_{dz}]$,
$v_d \in [0.5, 1.0]$, and
$r_a \in [R_f/2,R_f]$. We take the attacker velocity to be 
$v_a = 1.0$.

\begin{figure}
\centering
\includegraphics[width=3.2in]{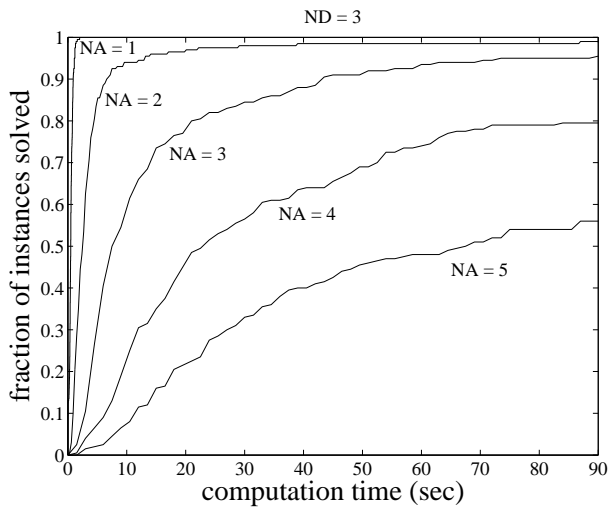}
\caption{Fraction of instances solved versus
computation time. The cost function is the number of attackers that
enter the Defense Zone plus a penalty on control effort
(equation~(\ref{eqn:objective}) with $\epsilon>0$). We consider
instances with defender set of size three ($N_D=3$) and attacker sets
of size $N_A=1,2,3,4,5$. To generate each curve, 200 random instances
were solved.  The parameters are $M_u=4$, $M_I=4$, $M_{dz}=4$,
$N_u=15$, and $N_a=15$. 
}
\label{acomplexity}
\end{figure}

\begin{figure}
\centering
\includegraphics[width=3.2in]{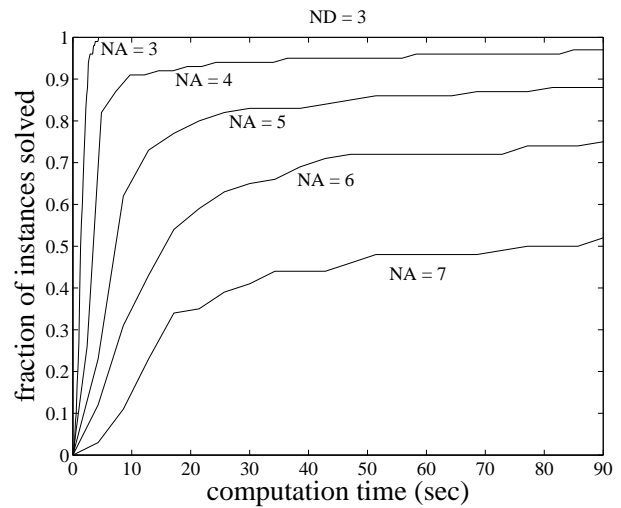}
\caption{Fraction of instances solved versus
computation time. The cost function is the number of
attackers that enter the Defense Zone (equation~(\ref{eqn:objective})
with $\epsilon=0$). We consider 
instances with defender set of size three ($N_D=3$) and attacker
sets of size $N_A=3,4,5,6,7$. To generate each curve, 200 random
instances were solved.
The parameters are
$M_u=4$, $M_I=4$, $M_{dz}=4$, $N_u=15$, and
$N_a=15$.} 
\label{acomplexity2}
\end{figure}

In Figure~\ref{acomplexity}, we plot the fraction of instances solved
versus computation time for the case where the cost function includes
the number of attackers that enter the Defense Zone plus a penalty on
the control effort. This cost function is given by
equation~(\ref{eqn:objective}) with $\epsilon = 0.1$.  In
Figure~\ref{acomplexity2}, we plot the fraction of instances solved
versus computation time for the case where the cost function includes
only the number of attackers that enter the Defense Zone.  This cost
function is given by equation~(\ref{eqn:objective}) with $\epsilon =
0$.

As these figures show, the case where the cost function only includes
the number of attackers that enter the Defense Zone is less
computationally intensive than the case where the cost function also
includes a penalty on the control effort.  For example, for the case
where $\epsilon = 0$, $N_D = 3$, and $N_A = 4$, 50\% of the problems
are solved in 3.5 seconds.  However, for the case where $\epsilon =
0.1$, $N_D = 3$, and $N_A = 4$, 50\% of the problems are solved in 78
seconds.  

When $\epsilon = 0$ in the cost function, the solver stops once a set
of trajectories for the defenders is found that results in the minimum
number of attackers entering the Defense Zone. This trajectory is
often not the most efficient trajectory with respect to the control
effort of the defenders. For the case where $\epsilon = 0.1$, once a
set of defender trajectories is found that results in the minimum
number of attackers entering the Defense Zone the solver continues to
search. The solver searches until it finds a set of defender
trajectories that not only results in the minimum number of attackers
entering the Defense Zone but also uses the defender control effort in
the most efficient way.

\begin{figure}
\centering
\includegraphics[width=3.2in]{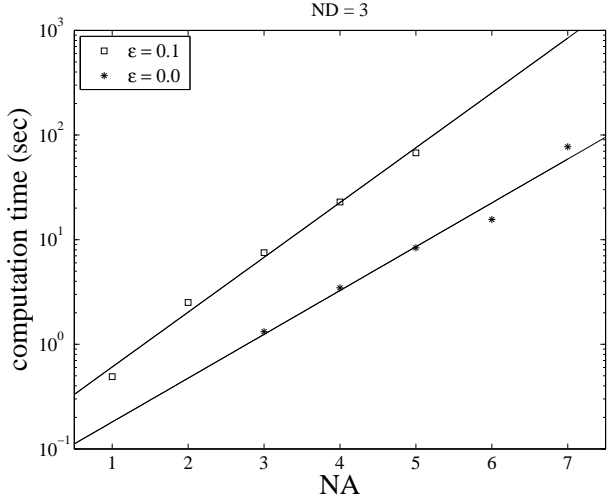}
\caption{Computation time necessary to solve 50\% of the randomly
generated instances versus the number of attackers. For all instances,
the defender set is of size three ($N_D=3$). Each square denotes a
data point for the case where the cost function is the number of
attackers that enter the Defense Zone plus a penalty on each
defender's control effort ($\epsilon = 0.1$ case).
Each asterisk denotes a
data point for the case where the cost function is the number of
attackers that enter the Defense Zone ($\epsilon = 0$ case).
The solid lines denote fitted exponential functions to these data.
} 
\label{expGrowth}
\end{figure}

In Figure~\ref{expGrowth}, we plot the computation time necessary to
solve 50\% of the randomly generated instances versus the size of the
attacker set. For all instances considered, the defender set is of
constant size ($N_D=3$). We plot the data for both cost functions
considered above ($\epsilon = 0$ and $\epsilon = 0.1$). This figure
shows that the computation time grows exponentially with the number of
vehicles in the attacker set. 


\section{Discussion}
\label{sec:discussion}
In this paper, we showed how to use MILPs to model and generate
strategies for multi-vehicle control problems.  The MILP formulation
is very expressive and allows all aspects of the problem to be taken
into account. This includes the dynamic equations governing the
vehicles, the dynamic equations governing the environment,  the state
machine governing adversary intelligence, logical constraints, and
inequality constraints. The solution to the resulting MILP is the
optimal team strategy for the problem.  As shown by our average case
complexity study, the MILP method becomes computational burdensome for
large problems and thus, for these cases, is not suitable for real-time
applications.

There are several steps that can be taken to make the MILP approach
applicable for real-time multi-vehicle control applications.  One
approach is to solve the MILPs faster.  This can be done by using a
more powerful computer or by distributing the computation over a set
of processors. For the multi-vehicle control problems, it may be
advantageous to distribute the computation over the set of vehicles.
Distributed methods for solving MILPs have been considered
in~\cite{Ralphs03}.  

Another approach is to move all, or some components of, the computational
burden offline. This can be done by formulating the problem as a
multi-parametric MILP.  A multi-parametric MILP is a MILP formulated
using a set of parameters each allowed to take on values over an
interval.  
This problem is much more computationally intensive than the MILP
problems considered in this paper. However, because the computation is
performed offline, this is not an issue unless the computation takes
an unreasonable amount of time. With the solution to the
multi-parametric MILP, a table can be formed and used to look up
optimal team strategies for the multi-vehicle system in real time.
Multi-parametric MILP control  problems can be solved using the 
Multi-Parametric Toolbox~\cite{mpt}.

Finally, we discuss efficient MILP formulations as a way of decreasing
computational requirements.  The computational effort required to
solve a MILP depends most on the number of binary variables used in
its MILP problem formulation. Thus, reducing the number of binary
variables is advantageous.  In this paper, our motivating problem
required three different time discretizations.  The discretizations
included one for the control input to the vehicles, one for obstacle
avoidance, and one for attacker intercept.  With each discretization
step, a set of binary variables must be added to the MILP formulation.
In most of the instances solved in this paper, we discretized time
uniformly. However, we posed the MILP in a general way such that
nonuniform time discretizations could be used. This allows for
intelligent time distribution selection, which can significantly
reduce the required computational effort to solve a problem.
In~\cite{Earl04b}, we developed several iterative MILP techniques for
intelligent discretization step selection. 

\appendices
\section{}
\label{logicconvert}
In this appendix, we illustrate how to convert a logic expression into
an equivalent set of inequalities using the procedure from~\cite{tyler}.
First the logic expression is converted into a conjunction of
disjunctions, called conjunctive normal form (CNF), by applying
tautologies.  For example, let the variables $p_i \in \{0,1\}$ 
be binary variables. The expression
\begin{eqnarray}
\label{lc1}
&&\mbox{$((p_1 = 1)$ or $(p_2 = 1))$}\nonumber\\
&&\mbox{and $((p_3 = 0)$ or $(p_2 = 1))$}\nonumber\\
&&\mbox{and $((p_4 = 1)$ or $(p_5 = 1)$ or $(p_6 = 0))$}
\end{eqnarray}
is in CNF. 

Once we have converted the logic expression into CNF we can easily
write each disjunction as an inequality constraint that must be
satisfied in order for the disjunction to be true.  For example, the
second disjunction of equation~(\ref{lc1}), $((p_3=0)$ or $(p_2=1))$, 
is true if and only if $(1-p_3) + p_2 \geq 1$.  Therefore,
equation~\ref{lc1} is true if and only if the following inequalities
hold
\begin{eqnarray}
  &&p_1 + p_2 \geq 1\nonumber\\
  &&(1-p_3) + p_2 \geq 1\nonumber\\
  &&p_4 + p_5 + (1-p_6) \geq 1.
\end{eqnarray}

\subsection{Equation~(\ref{gamma2})}
First consider the ($\Rightarrow$) direction of equation~(\ref{gamma2}).
Replacing implication with disjunction we have
\begin{eqnarray}
  &&(\gamma[k_a] = 1) \mbox{ or }\nonumber\\ 
  &&(g_m[k_a] = 1\;\;\;\;
  \forall m \in \{1,\ldots,M_{dz} \})
\end{eqnarray}
and distributing the OR we have 
\begin{eqnarray}
  \left(
  \gamma[k_a] = 0 \mbox{ or } g_m[k_a] = 1 
  \right)\;\;\;\;
  \forall m \in \{ 1,\ldots,M_{dz}\}
\end{eqnarray}
which is equivalent to 
\begin{eqnarray}
  &&\left( \gamma[k_a] = 0 \mbox{ or } g_1[k_a] = 1 \right)\nonumber\\
  &&\mbox{and }\left( \gamma[k_a] = 0 
    \mbox{ or } g_2[k_a] = 1 \right)\nonumber\\
  &&\vdots\nonumber\\
  &&\mbox{and }\left( \gamma[k_a] = 0 \mbox{ or } g_{M_{dz}}[k_a] = 1
    \right).
\end{eqnarray}
This expression is in CNF, therefore it is equivalent to the following
set of inequality constraints
\begin{eqnarray}
    (1-\gamma[k_a]) + g_m[k_a]  \geq 1\;\;\;\; 
    \forall m \in \{ 1,\ldots,M_{dz} \}. 
\end{eqnarray}
Now consider the ($\Leftarrow$) direction of equation~(\ref{gamma2}).
Replacing implication with disjunction 
\begin{eqnarray}
  &&(\gamma[k_a] = 1) \mbox{ or }\nonumber\\ 
  &&\neg (g_m[k_a] = 1\;\;\;\;\forall m \in \{1,\ldots,M_{dz} \})
\end{eqnarray}
and distributing the negation we have
\begin{eqnarray}
  \left( \gamma[k_a] = 1 \right) 
  \mbox{ or } 
  \left( 
  \exists m \mbox{ such that }
  g_m[k_a] = 0 
  \right).
\end{eqnarray}
which is equivalent to
\begin{eqnarray}
  &&\left( \gamma[k_a] = 1 \right)\nonumber\\
  &&\mbox{or } \left( g_1[k_a] = 0 \right)\nonumber\\
  &&\mbox{or } \left( g_2[k_a] = 0 \right)\nonumber\\
  &&\vdots\nonumber\\
  &&\mbox{or } \left( g_{M_{dz}}[k_a] = 0 \right).
\end{eqnarray}
This expression, which is a disjunction, is in CNF and therefore 
is equivalent to the following inequality
\begin{eqnarray}
   \sum_{i=1}^{M_{dz}}(1-g_i[k_a]) + \gamma[k_a] \geq 1.
\end{eqnarray}
\subsection{Equation~(\ref{asgovlogic})}
First consider the $(\Rightarrow)$ direction of equation~(\ref{asgovlogic}).
Replacing implication with the equivalent disjunction we have
\begin{eqnarray}
  &&(a[k_a+1] = 0) \mbox{ or } \nonumber\\
  &&\left( a[k_a]=1 \mbox{ and } \gamma[k_a]=0 \mbox{ and } 
  \delta[k_a]=0 
  \right) 
\end{eqnarray}
and distributing OR over AND we have
\begin{eqnarray}
  &&\left( a[k_a+1] = 0 \mbox{ or }  \delta[k_a]=0 \right)\nonumber\\
  &&\mbox{and }\left( a[k_a+1] = 0 \mbox{ or }  a[k_a]=1 \right)%
  \nonumber\\
  &&\mbox{and }\left( a[k_a+1] = 0 \mbox{ or }  \gamma[k_a]=0 \right).
\end{eqnarray}
Now that the expression is in CNF we can easily write
it as a set of inequality constraints.
\begin{eqnarray}
  &&( 1-a[k_a+1] ) + \left( 1-\delta[k_a] \right) \geq 1\nonumber\\
  &&\left( 1-a[k_a+1) \right) + a[k_a] \geq 1\nonumber\\
  &&\left( 1-a[k_a+1) \right) + \left( 1-\gamma[k_a] \right) \geq 1.
  \label{asineq1}
\end{eqnarray}
Now consider the other direction $(\Leftarrow)$ of
equation~(\ref{asgovlogic}). 
Replacing the implication with disjunction we have
\begin{eqnarray}
  &&( a[k_a+1] = 1 ) \mbox{ or }\nonumber\\ 
  &&\neg \left( \delta[k_a] = 0 \mbox{ and } a[k_a] = 1 \mbox{ and } 
  \gamma[k_a] = 0 \right) 
\end{eqnarray}
and distributing the negation we have
\begin{eqnarray}
  \delta[k_a] = 1 \mbox{ or } a[k_a] = 0 \mbox{ or } \gamma[k_a] = 1 
  \mbox{ or } a[k_a+1] = 1
\end{eqnarray}
this disjunction is equivalent to the flowing inequality 
\begin{eqnarray}
  \delta[k_a] + (1-a[k_a]) + \gamma[k_a] + a[k_a+1] \geq 1.
  \label{asineq2}
\end{eqnarray}
In summary, we have taken the governing equations for the attacker's 
binary state~(\ref{asgov2}) and derived an equivalent set of
inequalities:~(\ref{asineq1}) and~(\ref{asineq2})
which can be simplified to the inequalities given by
equation~(\ref{attdyn3}).

\section*{Acknowledgment}
We thank J.~Ousingsawat for helpful comments and encouragement. 


\end{document}